\documentclass[journal]{IEEEtran}

\usepackage{amsmath,amsfonts}
\usepackage{algorithm2e}
\usepackage{array}
\usepackage{textcomp}
\usepackage{stfloats}
\usepackage{url}
\usepackage{verbatim}
\usepackage{graphicx}
\usepackage{adjustbox}
\usepackage{cite}
\usepackage{hyperref}
\usepackage{url}
\usepackage{booktabs}
\usepackage{braket} 
\usepackage{soul}
\usepackage{placeins}
\usepackage[dvipsnames]{xcolor}
\usepackage{colortbl} 
\usepackage{caption, subcaption}
\SetKwComment{Comment}{/* }{ */}
\RestyleAlgo{ruled}

\usepackage{xcolor} 

\usepackage[switch]{lineno}
\usepackage{multicol}
\usepackage{multirow}
\usepackage{mwe}
\usepackage{float}
\usepackage{balance}
\usepackage{lipsum}
\usepackage{hyperref}

\ifCLASSINFOpdf
\else
\fi
\begin{document}

\title{A Quantum-assisted Attention U-Net for Building Segmentation over Tunis using Sentinel-1 Data}

\author{\IEEEauthorblockN{Luigi Russo, \IEEEmembership{Student Member, IEEE},
Francesco Mauro, \IEEEmembership{Student Member, IEEE}, \\
Babak Memar, \IEEEmembership{Student Member, IEEE}, 
Alessandro Sebastianelli, \IEEEmembership{Member, IEEE},\\
Silvia L. Ullo, \IEEEmembership{Senior Member, IEEE}, 
and Paolo Gamba, \IEEEmembership{Fellow, IEEE} 
}
\thanks{Luigi Russo and Paolo Gamba are with the Department of Electrical, Computer and Biomedical Engineering, University of Pavia, 27100 Pavia, Italy (e-mail: luigi.russo02@universitadipavia.it, paolo.gamba@unipv.it)\\
Francesco Mauro and Silvia Liberata Ullo are with the Department of Engineering, University of Sannio, 82100 Benevento, Italy (email: \{f.mauro@studenti., ullo\}@unisannio.it).\\ 
Babak Memar is with the Department of Civil, Building and Environmental Engineering, Sapienza University of Rome, 00184 Rome, Italy (e-mail: babak.memar@uniroma1.it).
Alessandro Sebastianelli is with the $\mathsf{\Phi}$-lab, European Space Agency, Frascati, Italy (email: Alessandro.Sebastianelli@esa.int)}}

\markboth{Joint Urban Remote Sensing Event (JURSE) 2025}{}
\maketitle
\begin{abstract}
Building segmentation in urban areas is essential in fields such as urban planning, disaster response, and population mapping. Yet accurately segmenting buildings in dense urban regions presents challenges due to the large size and high resolution of satellite images. 
This study investigates the use of a Quanvolutional pre-processing to enhance the capability of the Attention U-Net model in the building segmentation. Specifically, this paper focuses on the urban landscape of Tunis, utilizing Sentinel-1 Synthetic Aperture Radar (SAR) imagery.

In this work, Quanvolution was used to extract more informative feature maps that capture essential structural details in radar imagery, proving beneficial for accurate building segmentation.
Preliminary results indicèate that proposed methodology achieves comparable test accuracy to the standard Attention U-Net  model while significantly reducing network parameters. This result aligns with findings from previous works, confirming that Quanvolution not only maintains model accuracy but also increases computational efficiency. These promising outcomes highlight the potential of quantum-assisted Deep Learning frameworks for large-scale building segmentation in urban environments.
\end{abstract}

\begin{IEEEkeywords}
CNN, U-Net, neural network, Deep Learning, SAR, Sentinel-1, Quantum, Quanvolution.
\end{IEEEkeywords}

\IEEEpeerreviewmaketitle

\section{Introduction}
\IEEEPARstart{B}{uilding} segmentation in urban areas is crucial for applications like urban planning, disaster management, and population tracking. With continuous urbanization and advancements in satellite imaging, the demand for accurate segmentation methods has significantly grown. However, identifying buildings in densely built environments remains challenging due to complex urban structures and spatial resolution constraints \cite{s19020333}. Buildings exhibit diverse characteristics in color, shape, and roof structure, with variations in size and form based on intended use, contributing to their geometric complexity \cite{8444434, AMIRGAN2024101176}. Several studies in the state of the art (SOTA) focus on performing 2D analysis of urban areas, aiming to segment buildings and extract useful information from high-resolution imagery. For example, Chen et al. \cite{CHEN2023129} combined color normalization, super-resolution, scene classification, building extraction, and mosaicking to extract buildings from super-resolution images in Japan. Similarly, \cite{rs15092347} applied Sentinel-2 data and super-resolution to improve segmentation.
In recent years, Deep Learning (DL) models, especially U-Net \cite{ronneberger2015unetconvolutionalnetworksbiomedical}, have shown great promise in capturing spatial details for segmentation tasks. Many studies have leveraged DL techniques in remote sensing to identify and delineate buildings in urban environments. For example, Memar et al. \cite{memar2024u} applied a U-Net model to Very High Resolution (VHR) imagery from COSMO-SkyMed (CSK), while an encoder-decoder network with selective spatial pyramid dilation (SSPD) and context balancing modules (CBM) improved SAR-based building detection \cite{jing2021fine}. Sikdar et al. proposed a Fully Complex-valued Fully Convolutional Multi-feature Fusion Network (FC2MFN) incorporating SAR phase information \cite{sikdar2022fully}. A dual-attention U-Net (DAU-Net) was introduced for SAR-based monitoring of sea ice and open water \cite{ren2021development}, and DeepMAO, a dual-branch network combining SAR and EO imagery, improved segmentation in dense urban areas \cite{sikdar2023deepmao}. Additionally, change detection methods applied to KH-9 and Sentinel-2 images tracked urban sprawl in Bishkek, aiding in seismic risk assessment \cite{watson2022analyzing}.

Building on these advancements, this work focuses on the use of an Attention U-Net model for building segmentation in the urban area of Tunis, utilizing Sentinel-1 (S1) SAR data. The key advantage of SAR lies in its ability to capture detailed structural information in complex urban environments, even under cloud cover, and is further enhanced by the integration of Quantum Computing (QC) techniques.

In particular, QC in Earth Observation (EO) for urban remote sensing remains underexplored in the SOTA. Existing research combines quantum and classical techniques to reduce the computational load of complex models. For example, Fan et al. \cite{fan2023urban} introduced hybrid quantum-classical frameworks—MQCNN and FQCNN—for urban land cover classification using S2 imagery, where QC is employed for feature extraction and classical layers for final classification. Their models achieved promising accuracy, demonstrating QC's potential in EO for feature extraction and classification. However, the quantum convolution layer proposed in \cite{fan2023urban} for feature extraction was less efficient than the \textit{quanvolution} operator proposed by Sebastianelli et al. \cite{sebastianelli2024quanv4eo, mauro2024qspecklefilter}, which requires fewer qubits.

Building upon this, in the present work, we apply the \textit{quanvolution} operator to process Sentinel-1 (S1) data from the urban area of Tunis, and use the processed data as input for pixel-wise classification through the Attention U-Net model.

\label{intro}

\begin{figure*}[!ht]
    \centering
    \includegraphics[width=0.9\columnwidth]{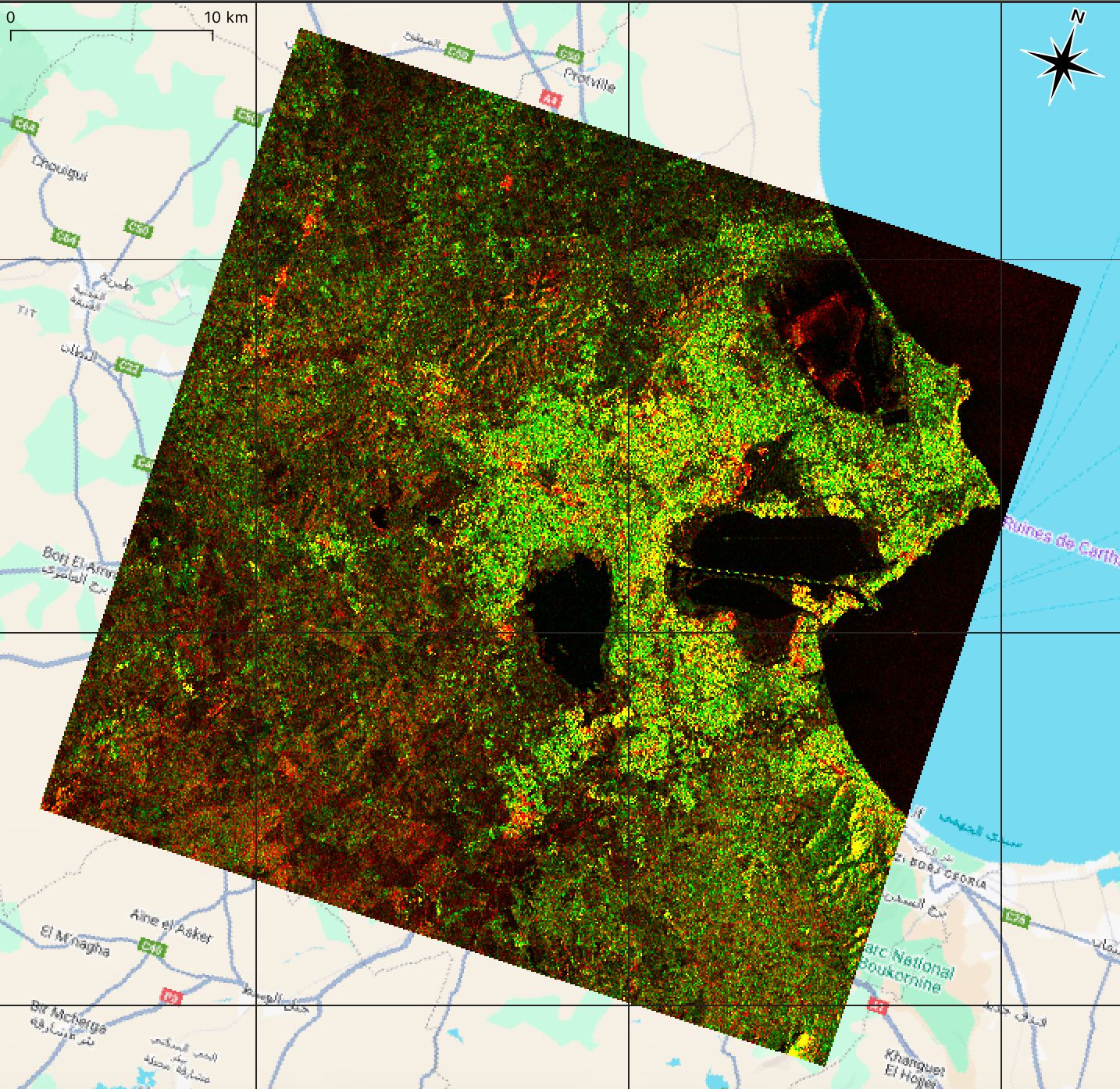}
    \caption{Input study area depicting the entire urban area of Tunis, acquired through Sentinel-1 SAR data.}
    \label{map_input}
\end{figure*}

\section{Input data and study area}
For training the proposed model, we downloaded a S1 image (\href{https://sentinels.copernicus.eu/web/sentinel/user-guides/sentinel-1-sar/revisit-and-coverage#:~:text=A%20single%20SENTINEL%2D1%20satellite,repeat%20cycle%20at%20the%20equator.}{ESA: S1}) covering the urban area of Tunis (with a polygon of $\sim 50 km^2$), the capital city of Tunisia, located in North Africa along the Mediterranean coast, as shown in Figure \ref{map_input}. Data were retrieved from the \textit{S1 SAR GRD collection} on Google Earth Engine (GEE), containing both VV and VH polarizations.

As ground truth (GT), we utilized building polygons from the \href{https://developers.google.com/earth-engine/datasets/catalog/GOOGLE_Research_open-buildings_v3_polygons#table-schema}{Open Buildings $2.5$D Temporal dataset}, which includes over $1.8$ billion building footprints across Africa, Latin America, the Caribbean, South Asia, and Southeast Asia. The dataset offers data on building presence, counts, and heights at a $4$m resolution, updated annually from $2016$ to $2023$.

To prepare the training and test datasets, we performed a patching operation on both the input image and the corresponding GT. The original image was divided into $256\times256$ pixel patches with a factor of stride equal to $128$, generating partially overlapping \textit{chips} to augment the dataset. 
The resulting patches were split into training and testing datasets, with the training set used for model training and the test set for evaluation.

\section{Methodology}\label{sec:methods}

\begin{figure}[htp]
    \centering
    \includegraphics[width=0.6\columnwidth]{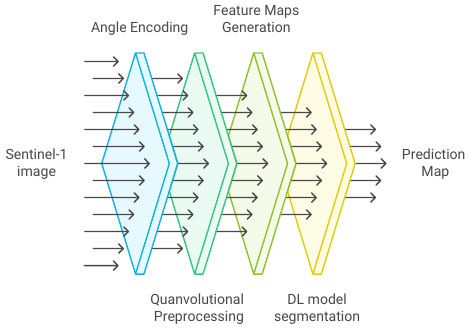}
    \caption{Diagram of the proposed methodology, illustrating SAR data preprocessing through a quanvolutional operator with angle encoding to enhance feature extraction, followed by the DL framework for urban area segmentation.}
    \label{schematic}
\end{figure}

The following section describes the proposed methodology, $1)$ it presents the preprocessing of SAR data with a quanvolution operator, aimed at increasing the number of extractable feature maps from the data, $2)$ it introduces the Deep Learning (DL) framework used to segment the urban area. The overall workflow is illustrated in Figure \ref{schematic}.

The sequence of steps/blocks is described as follows.\\
\paragraph{Quanvolutional Pre-Processing}
a novel preprocessing methodology for urban classification of S1 data is introduced, utilizing a quanvolutional operator, as proposed by Sebastianelli et al. \cite{sebastianelli2024quanv4eo}.
To represent classical data ${\mathbf{x} \in \mathrm{R}^n}$ within an $n$-qubit quantum circuit, a quantum feature map $\phi: \mathrm{R}^n \rightarrow H^{2^n}$ is used, where $H^{2^n}$ is a Hilbert space of dimension $2^n$. This involves applying a unitary operator $U_{\phi}(\mathbf{x})$ to the initial state $\ket{0}^{\otimes n}$, resulting in:

\begin{equation} 
    U_{\phi}(\mathbf{x})\ket{0}^{\otimes n} = \ket{\phi(\mathbf{x})} = \ket{\psi}.
\end{equation}

In the proposed framework, each patch of the input image is encoded into a quantum state using \textbf{angle encoding} \cite{weigold2020data}. In this way,  the input data are transformed into a representation compatible with quantum processing.

Following angle encoding, we implement a quanvolutional operator applied to quantum circuits  to perform convolutional operations. This layer leverages the inherent parallelism offered by quantum mechanics, allowing for efficient processing of the data. Notably, the parameters within the quanvolutional operator are frozen, meaning that they are not subject to training during the optimization process, as propsoed by Sebastianelli \textit{et Al.} \cite{sebastianelli2024quanv4eo}.


As the quanvolutional operator processes the encoded input, it extracts more informative feature maps, effectively capturing a wide range of spatial features present in the data.

The configuration of the layer is influenced by several critical parameters. The number of \textit{qubits} plays a pivotal role in determining the quantum representation of the input data and must satisfy the constraint:

\begin{equation}
    \centering
    qubits \geq kernel\ size^2
\end{equation}

This condition ensures that there are sufficient qubits to effectively represent the convolutional kernel, which directly impacts the model's ability to extract relevant features. The \textit{kernel size} defines the spatial dimensions of the convolutional kernel and influences the receptive field during the convolutional operations.

After processing through the quanvolutional operator, a measurement stage is employed to convert the quantum states back to the classical domain. This crucial step allows the derived features to be utilized in subsequent classification tasks while preserving the information captured during the quantum processing phase. Upon exiting this quanvolutional operator, the new feature maps are given as input to the classical Attention U-Net model.

\paragraph{Deep Learning framework}

The Deep Learning model used to segment buildings from input S1 data is the \textit{Attention U-Net}. This framework, introduced by \cite{oktay2018attentionunetlearninglook}, leverages attention mechanisms to improve the extraction of relevant features. The architecture builds on the conventional U-Net \cite{ronneberger2015unetconvolutionalnetworksbiomedical}, consisting of a contracting path followed by an expansive path. This structure enables the model to capture both broad context and detailed information, which is particularly useful in urban image segmentation tasks.

The enhancement in this architecture comes from the incorporation of an \textit{attention gate}, which accentuates pertinent regions while diminishing feature activations in less relevant areas. This block is integrated as a \textit{skip connection} in the overall framework, proving particularly useful for identifying built-up areas in satellite images. It allows the model to focus on distinctive building patterns and shapes, resulting in more precise segmentations.

\begin{equation}
\tilde{g} = W_g \ast g
\label{eqn1}
\end{equation}

\begin{equation}
\tilde{x}^i = W_x \ast x^i
\label{eqn2}
\end{equation}

\begin{equation}
\psi = \tilde{g} + \tilde{x}^i
\label{eqn3}
\end{equation}

\begin{equation}
\psi' = \text{ReLU}(\psi)
\label{eqn4}
\end{equation}

\begin{equation}
\psi'' = \text{BatchNorm}(\psi')
\label{eqn5}
\end{equation}

\begin{equation}
\alpha = \sigma_2(\psi'')
\label{eqn6}
\end{equation}

\begin{equation}
\rho = W_\rho \ast \alpha
\label{eqn7}
\end{equation}

\begin{equation}
x_{\text{out}}^i = \rho \cdot x^i
\label{eqn8}
\end{equation}

\begin{figure}[htp]
	\centering
	\includegraphics[width=0.8\columnwidth]{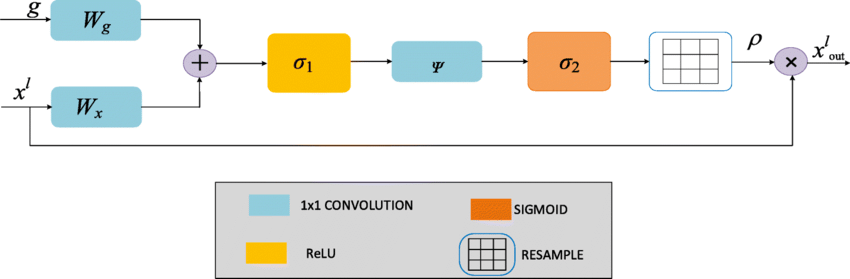}
	\caption{Workflow of the \textit{attention gate} used to enhance the performance in 2D building segmentation tasks.}
	\label{attention_gate}
\end{figure}

As illustrated in Fig. \ref{attention_gate}, the attention block takes two inputs: \( g \) and \( x^i \). Here, \( g \) represents features from the decoder (expanded features), and \( x^i \) represents features from the encoder (compressed features) at a particular depth in the network. Both feature sets undergo a \( 1 \times 1 \) convolution, with the decoder features \( g \) being convolved with learnable weights \( W_g \) (see \ref{eqn1}), and the encoder features \( x^i \) convolved with learnable weights \( W_x \) (\ref{eqn2}).

These results are then summed element-wise (\ref{eqn3}), and passed through a ReLU activation function (\ref{eqn4}). Next, Batch Normalization is applied to stabilize the learning process (\ref{eqn5}).

The output of the BatchNorm layer is then passed through another \( 1 \times 1 \) convolution followed by a Sigmoid activation function, which generates the attention map \( \alpha \) (\ref{eqn6}). This attention map is subsequently resampled through a \( 1 \times 1 \) convolution operation (\ref{eqn7}) to ensure spatial alignment with the input features.

Finally, the resampled attention map \( \rho \) is multiplied element-wise with the original encoded features \( x^i \) (\ref{eqn8}), producing the output \( x_{\text{out}}^i \), which highlights the important regions in the input features for the segmentation task.

\begin{table*}[htp]
    \centering
    \caption{Performance comparison between the standard Attention U-Net and the Quantum-assisted Attention U-Net using different quantum configurations.}

    \label{tab:full_results}
    \resizebox{0.9\textwidth}{!}{
        \begin{tabular}{llcccccc}
            \toprule
            \textbf{Model Type} & \textbf{Circuit} & \textbf{Overall Accuracy (O.A.)} & \textbf{Trainable} & \textbf{Qubits} & \textbf{Layers} & \textbf{Filters} & \textbf{Kernel} \\
            ~ & ~ & \textbf{(O.A.)} & \textbf{ Parameters} & \textbf{(n\_qubits)} & \textbf{(n\_layers)} & \textbf{(num\_filters)} & \textbf{Size} \\
            \midrule
            Attention U-Net & - & 0.9491 & 34.8 million & - & - & - & - \\
            Quanvolution + Attention U-Net & Strongly Entangled Circuit & 0.9373 & 2.1 million & 9 & 2 & 9 & 3x3 \\ 
            Quanvolution + Attention U-Net & Random Circuit & 0.9343 & 2.1 million & 9 & 2 & 9 & 3x3 \\ 
            Quanvolution + Attention U-Net & Basic Entangled Circuit & \textbf{0.9384} & 2.1 million & 9 & 2 & 9 & 3x3 \\
            \bottomrule
        \end{tabular}
    }
    \label{table_results}
\end{table*}

\section{Results and Discussion} \label{sec:results}
The results, summarized in \textbf{Table \ref{table_results}}, indicate that integrating quanvolutional operators with the Attention U-Net architecture significantly reduces the number of trainable parameters while maintaining competitive performance in terms of overall accuracy (OA).

\begin{figure} [htp]
    \centering
    \includegraphics[width=0.9\columnwidth]{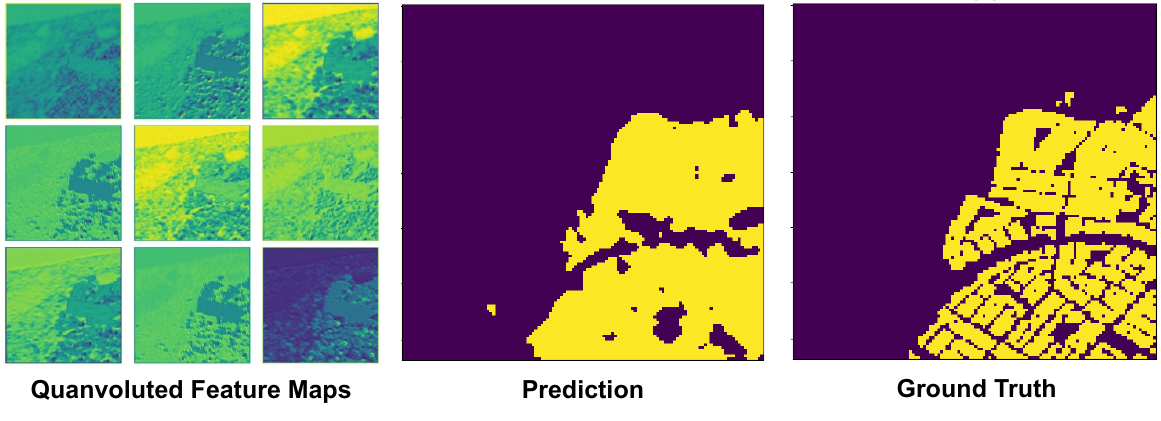}
    \caption{Comparison of quanvoluted feature maps, prediction, and corresponding ground truth for a random sample from the test set.}
    \label{Featuremaps_PR_GT}
\end{figure}

The classic Attention U-Net achieves the highest OA of 0.9491, but it requires 34.8 million parameters, which increases computational demands. In contrast, the quantum-assisted models achieve OA values between 0.9343 and 0.9384 with only 2.1 million parameters—a reduction of over $93\%$. An example of the quanvoluted feature maps and the corresponding segmentation prediction, compared with the GT on the test set, is shown in Fig. \ref{Featuremaps_PR_GT}.

Three different circuit configurations were evaluated for the Quanvolution-enhanced models: \textit{Strongly Entangled Circuit}, \textit{Random Circuit} and \textit{Basic Entangled Circuit}. This last circuit reached the highest OA among the quantum models at 0.9384, approaching the classic Attention U-Net’s performance with significantly fewer trainable parameters.

\vspace{-10pt} 
\section{Conclusions}
This study explores the integration of quantum-based techniques, specifically Quanvolution, with Attention U-Net   for building segmentation in urban environments using Sentinel-1 SAR data. While the classic Attention U-Net   achieved the highest overall accuracy (OA) of 0.9491, it requires 34.8 million trainable parameters, which is computationally demanding. In contrast, the quantum-augmented models, regardless of the circuit configuration (strongly, random, or basic), achieve OAs ranging from 0.9343 to 0.9384, with a significant reduction in trainable parameters to 2.1 million. These results indicate that Quanvolution offers a promising trade-off, maintaining competitive performance while drastically reducing computational complexity. This makes Quanvolution an attractive solution for large-scale urban segmentation tasks.

Future work will focus on three primary directions: first, refining the existing models and exploring the potential of trainable quanvolutional operators to enhance feature extraction and segmentation capabilities; second, applying a trainable quanvolutional operator; 
and third, conducting a more in-depth analysis by comparing the results obtained using quanvoluted feature maps with those derived from backscatter intensity information combined with polarimetric descriptors. While our study relies on Sentinel-1 data, which provides dual-polarization (VV and VH) information, a fully polarimetric dataset could offer deeper insights into urban structures by leveraging different scattering mechanisms. However, as Sentinel-1 GRD data does not include fully polarimetric SAR information, such an analysis would require alternative data sources. Investigating the impact of fully polarimetric SAR data on building segmentation presents an interesting avenue for future research, particularly in assessing how quanvoluted feature maps can complement or enhance the use of polarimetric descriptors for a more detailed urban structure analysis.

\label{sec:conclusions}

\bibliographystyle{IEEEtran.bst}
\bibliography{main}

\end{document}